# Empirical learning aided by weak domain knowledge in the form of feature importance

**Ridwan Al Iqbal**

Department of Computer Science

American International University Bangladesh

Dhaka, Bangladesh

*Stopofeger@yahoo.com*

**Abstract**

Standard hybrid learners that use domain knowledge require stronger knowledge that is hard and expensive to acquire. However, weaker domain knowledge can benefit from prior knowledge while being cost effective. Weak knowledge in the form of feature relative importance (FRI) is presented and explained. Feature relative importance is a real valued approximation of a feature's importance provided by experts. Advantage of using this knowledge is demonstrated by IANN, a modified multilayer neural network algorithm. IANN is a very simple modification of standard neural network algorithm but attains significant performance gains. Experimental results in the field of molecular biology show higher performance over other empirical learning algorithms including standard backpropagation and support vector machines. IANN performance is even comparable to a theory refinement system KBANN that uses stronger domain knowledge. This shows Feature relative importance can improve performance of existing empirical learning algorithms significantly with minimal effort.

## 1    Introduction

Empirical learning methods are the dominant methods for supervised learning problems. But these methods are dependent on significant amount of training data and training time to perform well. Furthermore, It has been shown [1] that learning algorithms simply refine the knowledge provided through the inductive bias in the algorithm. So, a learner only learns what it already knows with a better accuracy. Therefore, providing more prior knowledge greatly improves performance. A learner's learned model will also usually be much more comprehensible if the learner takes existing knowledge into account [2].

There has been extensive research to combine prior knowledge into learning algorithms (e.g. TangentProp [3], FOCL [4], RAPTURE [5], KBANN [6], KSVM [7]). Different systems use different forms of prior knowledge. Many of them use propositional logic such as KBANN. Eccentric forms of prior knowledge have also been used such as, derivatives of instances in TangentProp [3]; Certainty factors in RAPTURE [5]. However, the use of these hybrid systems has not been prevalent in real world applications.

The main reason for this is the high cost and difficulty of obtaining domain knowledge [8] [9] . The prior knowledge needed in these systems is "deep" and extensive; in fact, they are called theory refinement systems; for they essentially refine an existing domain theory. Such domain theories are sufficient to classify instances on their own in many cases. This means much more resources and expertise are required to acquire such deep prior knowledge. But such deep prior knowledge may not be available or cost effective in many problem domains. However, we propose that incorporating a "weaker"/"shallow" form of domain knowledge may benefit learning by providing the best of both plain empirical and hybrid theory refinement systems.

Weaker knowledge-based learners should have the following capabilities: Firstly, they should be superior to plain empirical systems in terms of classification accuracy if given the same training set. Secondly, they should be equally accurate to the empirical systems with smaller training set or training time. Moreover, the domain knowledge should be acquirable with much less expertise and over wider problem types.

This paper introduces *Feature relative importance* (FRI) as a form of weaker/shallow prior knowledge. The idea is to weight features according to their importance or contribution in the classification. There are already feature selection and ranking algorithms available e.g. [10] [11] [12]. However, these algorithms are only preprocessors that select and modify a dataset, discarding irrelevant features. However, it cannot be concluded that the other dimensions are unimportant: a different (possibly even disjoint) subset of features may yield the same predictive accuracy. The actual learning algorithms do not take the importance of attributes into account after feature selection and consider all features to be equally important. Knowledge about importance can be used to guide through the search space and achieve faster and better learning.

*Feature relative importance* represents importance of features as a real valued normalized weight in the [0, 1] range. Human experts can provide feature importance much easier than stronger domain theories. So, knowledge acquisition will also be cheaper and faster.

We then present the **I**ANN (*Importance aided Neural Network*) algorithm to demonstrate the benefit of FRI in empirical learning and show its superior performance compared to ordinary Multilayer neural networks. IANN is a simple modification of the standard neural network learning procedure but attains much higher performance. Briefly, IANN algorithm uses FRI initialize the NN and then uses a modified form of Backpropagation to train the network.

In the subsequent section, IANN is applied in the real world domain of molecular biology, specifically the problem of recognizing *eukaryotic splice junctions* and *promoter genes*. The performance is analyzed and compared with many popular learning algorithms as well as hybrid learning systems. It is shown that IANN performs better than plain empirical systems such as neural networks, C4.5 and support vector machines. In fact, IANN performs comparable to KBANN which uses a much stronger prior knowledge about the domain.

Brief outline of the paper: Section 2 provides an explanation of FRI, section 3 describes the IANN while section 5 describes the experimental results. Section 5 concludes the paper.

We now describe the notation used in this paper. $I_k$, is used to refer the FRI of a feature $A_k$. $A$ represents the set of all features in the data set $S$ of a learning model $L$. Neural network weights are represented by $w_{ij}$ which is the weight of the connection going from unit $i$ to unit $j$. $\delta_j$ represents the error of unit $j$; $\alpha$, learning rate, $X_{ij}$, the input between an unit $i$ to unit $j$. $Layer_k$ or $L_k$ is the set of units of layer k, where $k$ is 1 for the first hidden layer; $Inputs(k)$ is the set of all units that is input to $k$. $U$ is the set of all network units.

## 2     Feature relative importance

### 2.1     Feature relative importance formalism

Feature importance is provided by experts in our algorithm. But we must define what we mean by importance before we can use that knowledge. Importance is defined in terms of a feature's influence on a trained learner. A trained learner can be represented by a real valued scoring function $\psi : \mathcal{X} \to \mathbb{R}$ where, $\mathcal{X}$ is the input vector. There are many measures of feature importance provided in the feature selection and ranking domain e.g. sensitivity analysis [13]; FIRM [14]; saliency [11]. It generally depends on the domain; different definitions may be more appropriate depending on one's goals. Our definition is based on [14].

**Definition 1:** *Feature importance* of a feature $I_k^*$, is a representation of the total dependency of the learned model $\psi$ on the feature $A_k \in A$ to make a decision on a particular instance.

$$I_k^* ( A_k , \psi) \propto d_k \qquad (1)$$

We have to define the term "dependency" that we just used to define Feature importance. The exact definition will vary based on the learning model. But a general definition is:

**Definition 2:** *Dependency* $d_k$ is the expected score of learner's score function y conditional only to the feature $A_k$ given a input vector X:

$$d_k = E[\, y(X) \,|\, A_k ]$$

FRI is just a normalized approximation of Feature importance made by human experts. So, we can now define FRI.

**Definition 3:** *Feature relative importance* $I_k$ is a normalized approximation of actual importance of a feature $A_k$ relative to feature set A on the learning problem L . Feature relative importance $I_k$ is defined as:

$$I_k ( A, A_k , L ): I_k \in [0,1]$$

We can deduce the following relationship:

$$I_k \propto I_k^* \qquad (2)$$

Thus from (1) & (2):

$$I_k \propto d_k \qquad (3)$$

One important aspect of importance is that, FRI is a ratio. So, the knowledge comes from the variation in importance, not the absolute value of importance itself. A feature set with all features having FRI value of 1 is equivalent to a feature set having FRI value of 0.2 on all features. As long as the ratio is maintained between highly important and unimportant features, the value of FRI can be set arbitrarily.

FRI is a weak knowledge because apart from importance, no straight forward relationship between features and the actual learned model can be deduced from FRI. Human experts can point out the importance of a feature more easily than other forms of prior knowledge. So, it has less costly knowledge acquisition step than other theory refinement systems.

An example problem about "suitable weather conditions for play" is presented below to explain how FRI is used to represent an expert's knowledge about a feature's importance.

## 2.2 Playing condition example problem

This example problem has 7 features: sky, temperature, humidity, wind, date, season and traffic. We determine the importance of these features through intuition.

Table 1: Playing conditions problem feature importance to FRI conversion

| Feature | Importance | FRI |
|---|---|---|
| Sky | Medium | 0.7 |
| Temperature | High | 1.0 |
| Humidity | High | 1.0 |
| Wind | Medium | 0.7 |
| Date | Irrelevant | 0 |
| Road traffic | Irrelevant | 0 |
| Season | Low | 0.3 |

As we can see in Table 1, the actual FRI values are a matter of choice. The features Sky and Wind have medium importance but they are not given the value of 0.5. Any alternate set of

FRI values can also be chosen as long as the difference is maintained between features. Important features should have highest values while less-important features should have lower values.

In the subsequent section we present the IANN algorithm that uses FRI domain knowledge.

## 3   The IANN algorithm

### 3.1   Introduction

For an empirical learner, domain knowledge is normally used to tailor the bias that is otherwise provided through randomization or experimentation. Feed forward neural networks are provided bias in mainly three ways: the network topology, initial weights and the training algorithm used [15]. Network topology is usually set through experimentation; training algorithm popularly used is Backpropagation while network weights are normally initialized randomly.

IANN modifies normal multilayer perceptrons (MLP) by modifying weight initialization step and the training rule. Currently, we have not been able to find a way to tailor network architecture with FRI that actually improves performance. Importance is weak domain knowledge so it may not provide any insight about network topology. As a result, topology still must be selected by users in the traditional ways and must be fully connected.

We want to modify MLP using FRI, but the question is, "How the importance of a feature affects a particular network?" To answer this question, dependency for neural networks is defined. We know from definition 2 that the dependency is the output of the scoring function conditional only to the particular feature. The output function [15] of a single hidden layer neural network is:

$$\psi(X) = \sigma\left[\sum_{j \in L_1} w_{jy}\, \sigma\left(\sum_{i \in A} w_{ij} x_i + \theta_j\right) + \theta_y\right] \quad (4)$$

Thus, for a neural network which is a real valued non-linear function, dependency corresponds to the partial derivative of output function $y_k$ with respect to $A_k$:

**Definition 3:** *Dependency $d_k$ of feature $A_k$ on a **neural network** N is the partial derivative of y with respect to feature value $x_k$ of a feature $A_k$.*

This computation of dependency is straight forward for a linear learner such as perceptron, as the dependency on a feature is simply its weight. However, as the number of layers increase, the dependency gets distributed into the weights of higher layers and finding a straight forward dependency becomes difficult. The derivative of (4) is:

$$\frac{\partial y}{\partial x_k} = \frac{\partial \sigma(\varphi_2(y))}{\partial \varphi_2(y)} \sum_{j \in L_1} w_{jy} w_{kj} \frac{\partial \sigma(\varphi_1(j))}{\partial \varphi_1(j)} \quad (5)$$

Where,

$$\varphi_1(j) = \sum_{i \in A} w_{ij} x_i + \theta_j \quad (6)$$

$$\varphi_2(y) = \sum_{j \in L_1} w_{jy}\, \varphi_1(j) + \theta_y \quad (7)$$

Thus, the derivative depends on the current input to the network as well as the network weights. It has been shown that the computation of the derivative is np-complete [11]. Trying to use this particular formulation to influence neural networks through FRI is not a plausible choice. We instead use an assumption that higher dependency on a feature means higher average absolute weight on the first hidden layer that is directly connected to the input. Thus from (3) & (5) we can come to this following intuition which forms the basis of our algorithm:

**Intuition 1:** *In a feed forward network, Feature relative Importance of a feature is proportional to its average absolute weight on a trained network.*

$$I_k \propto \frac{1}{|L_1|} \sum_{j \in L_1} |w_{kj}| \qquad (8)$$

Clearly, this intuition is not correct in several cases. The relationship between importance and weight gets complex as more layers are added and the neural network gets more expressive power. Moreover, weights are not necessarily a proper measure of importance. In fact, by multiplying any feature of the inputs by a positive scalar and dividing the associated weight by the same scalar, the importance of the corresponding feature can be changed arbitrarily. However, experimentation has shown that our intuition works on many cases. So, this is a viable assumption.

From this intuition we can suppose that feature with higher FRI value should have more total weight in the first network layer that is directly connected with inputs; as that is the only layer where features have a direct role. Therefore, the focus of IANN is this layer. The training rule and weight initialization step are changed in IANN from ordinary MLP. They are explained in the next subsections.

### 3.2     IANN training rule

Based on intuition 1, features with higher FRI values should not only have higher total weights, but they should also be more "active" while training; that is, these features should have more weight change while training. IANN provides more "chance" to more important layers to change weights by providing them with a higher learning rate.

The general Backpropagation training rule [16] is as follows:

$\Delta w_{ij} = \alpha \, \delta_j X_{ij} \, ; \; i \in Inputs_j, \, j \in U;$

IANN provides a learning rate proportional to the FRI value. This is done by multiplying FRI value into delta weight. The training rule changes only for the units of the first Layer.

$\Delta w_{ij} = \alpha \, \delta_j X_{ij} I_i \, ; \qquad i \in Inputs_j, \, j \in Layer_1 \, ;$

However, this rule can be extended to all units by setting $I_k$ to be 1 for all hidden and output layer units. This simplifies implementation of the training rule. The updated training rule becomes,

$\Delta w_{ij} = \alpha \, \delta_j X_{ij} I_i \, ; \qquad i \in Inputs_j, \, j \in U;$

Due to this rule, important features will converge to local optimum faster while unimportant features will take more time to change weights. However, as unimportant features have proportionally less weight in a trained network from our intuition; this means a smaller learning rate is sufficient for convergence. Moreover, weight initialization will also have higher probability to set smaller weights into unimportant features. So, they have to make fewer changes into their weights.

### 3.3     Weight initialization

Weight initialization in IANN is similar to normal MLP except in the first hidden layer; where it is done in such a way so that important features have larger weight while less important features have lower weight. This can be usually done by simply setting the weight to be FRI and perturbing with a random number. However, testing has shown this to be ineffective. This is due to the fact that, this generates a general pattern of weight in all hidden layer units; larger weights in all important unit weights, while smaller weights in all unimportant units. Intuition 1 says only the average absolute weight will be proportional to FRI not the weights of all units. The power of hidden layers comes from the variation in initial weights. Thus, this simple weight initialization method sacrifices variation and also performance. So, we developed a new method of weight initialization that is based on our intuition and also improved performance.

First, a random number of features are selected for each of the units of the first hidden layer. The connection between these features and the unit is set to the FRI value with a randomly set sign. Other feature weights of the unit are set to a random number just like ordinary Multilayer perceptrons, but the range is half of that of the selected features. So, if the range of value for the selected features is [-1,1], the range for other features is [-0.5,0.5]. This ensures higher weight for features with FRI greater than 0.5. The weights of other layers are set randomly just like normal neural networks. IANN only modifies the first layer weight initialization procedure.

Table 2: IANN weight initialization

**Goal:** Set the weights of connections of $Layer_1$ based on FRI

**Require:** FRI $I_1....I_n$

1. **For** unit j ∈ $Layer_1$ **do**
2.     c = random number [0, n]; select c number of features randomly into $F_s$.
3.     **For** k ∈ feature set A **do**
4.       **If** k ∈ $F_s$
5.         Set $w_{ij}$ to $I_k$×randomchoose(1,-1)
6.       **Else**
7.         Set $w_{ij}$ to random number[-0.5,0.5]
8.     **End for**
9. **End for**

This procedure ensures some features get the FRI value as their initial weight in each hidden layer units but not all. The variation of initial weights is maintained while important features do have higher average initial weights. Only few features get "promoted" or "demoted" based on FRI while others have random weights.

## 4    Experimentation with IANN

### 4.1    Introduction and setup

This section reports the experimental results of using IANN which attains significant higher performance while being a simple modification. This is due to the benefit of using weak prior knowledge. Two real world problems of DNA analysis [6] were experimented[1]. We compared the datasets with several popular learning algorithms: standard backpropagation MLP [16], C4.5 [17], k-nearest neighbour [18] and support vector machines [19][2]. We also compared IANN with the theory refine system KBANN [6].

The first dataset is the promoter recognition dataset. A promoter, which is a short DNA sequence that precedes a gene sequence, is to be distinguished from a nonpromoter. The input is a sequence of 57 nucleotides (one of A, T, G or C). The dataset has 936 instances with 236 positive and 702 negative examples.

The second dataset is about splice-junction determination. This is a 3 class problem; the task is to determine into which of the three categories the specified DNA sequence belongs: Exon/Intron borders (EI), Intron/Exon borders (IE) or neither. The input is also a DNA sequence with 60 nucleotides. The dataset had 1007 instances selected randomly from a population of 3190. The percentage split of the classes is 25% EI, 25% IE & 50% neither

---

[1] The datasets are available at ftp://ftp.cs.wisc.edu/machine-learning/shavlik-group/datasets/

[2] Weka 3.6.2 (www.cs.waikato.ac.nz/ml/weka/) implementations of the empirical learning algorithms were used in the experiments.

examples. We employed the 10-fold cross validation methodology for both datasets.

A prior knowledge of propositional rules is also present with both datasets. We constructed the FRI for the datasets using these rules as a basis to recognize the important features. The features that are antecedent in more number of rules are given higher importance. The FRI given to these important features are 0.9, 0.8 and so on based on importance; while features that are not mentioned in any of the rules are given a low FRI of 0.3. This importance distribution is also similar to what found by [20]. The topology for the network is a single 23 unit hidden layer for the promoter dataset and 24 units for the splice-junction dataset. This topology generalizes as well as, or better than, all others we tried during a search of topology space.

The ordinary backpropagation NN had the same network topology as IANN . KBANN used the propositional rules as the domain theory. For the SVM, linear (u*v) kernel was found to be the most effective. For both IANN and ordinary NN, the network was trained for 100 epochs and no validation set was used.

### 4.2 Results

Table 3: (a) promoter dataset performance;    (b) Splice-junction dataset performance

| Learner | Correctness (%) |
|---|---|
| IANN | 94.97 |
| Backpropagation | 93.45 |
| SVM | 89.74 |
| C4.5 | 89.95 |
| Near. N. (k=3) | 90.49 |
| KBANN | 93.70 |

| Learner | Correctness (%) |
|---|---|
| IANN | 94.83 |
| Backpropagation | 93.23 |
| SVM | 88.77 |
| C4.5 | 90.86 |
| Near. N. (k=20) | 87.28 |
| KBANN | 93.68 |

It is apparent from the results that IANN outperforms other empirical algorithms and unexpectedly, it even outperforms KBANN in both datasets. IANN outperforms standard backpropagation learner even though IANN is a very simple adjustment of the standard algorithm.

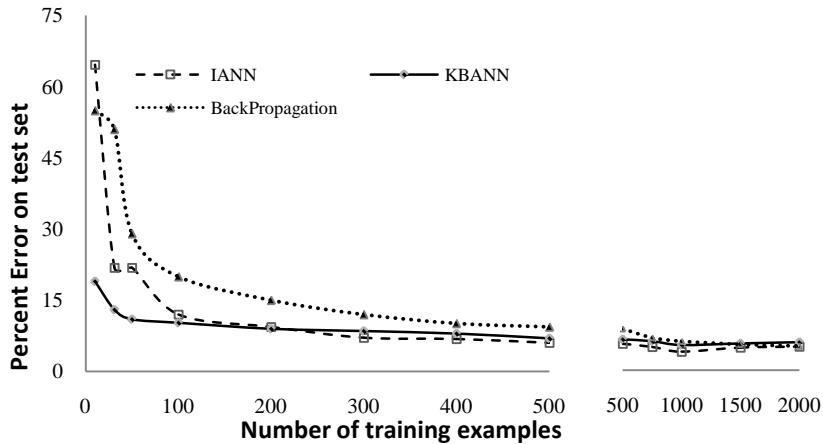

Figure 1: Learning curve for splice-junctions

The advantage of FRI prior knowledge can be more clearly evident by the splice-junctions learning curve in Figure 1. The splice junction set with full population of 3190 instances was used in this experiment. The training examples were randomly selected from the population while remaining instances became test examples. Error rate of ordinary backpropagation and KBANN were plotted for a comparison between the no knowledge, weak knowledge and strong knowledge learner. The results of KBANN and backpropagation were acquired from an earlier experiment [21].

We can see that KBANN learns the fastest with fewer examples in beginning. However, IANN quickly reaches KBANN within 100 training examples and continues to be better. An interesting trend is both knowledge aided algorithms reach an optimum performance and then the error rate actually climbs. But Backpropagation follows a decreasing trend throughout before reaching an error rate of 5.5%. IANN achieves the best performance of 4.3% for 1000 examples.

## 5       Conclusion and future directions

We have proposed a simple and efficient procedure of incorporating feature importance into neural network learning. The performance of such a learner shows feature importance aided learners can achieve superior performance over ordinary empirical learners and can even compare to stronger knowledge based learners without having the extra cost of a deep domain theory. This approach of incorporating feature importance into learners is worthy of further development. Possible future applications maybe areas where expert knowledge is not readily available and there is a scarcity of training data as well. IANN can be used in such domains with both fewer training data and expert knowledge. Furthermore, Modifications of existing popular empirical learners should also be developed that utilize feature importance. Currently it is assumed that feature importance knowledge provided by experts is almost correct to some extent. However, if its accuracy is questionable then performance will actually degrade. So, a learning algorithm can be developed that can correct FRI knowledge through training examples.

Current machine learning algorithms rely too much on training examples. Incorporating more and more domain knowledge is the way for improvement. Our proposed method can be a stepping stone towards that goal for domains where deep knowledge is not available or cost effective.